\documentclass[letterpaper, 10 pt, conference]{IEEEtran}  

\IEEEoverridecommandlockouts                              

\usepackage{amsmath}
\usepackage{amssymb}
\usepackage{siunitx}
\usepackage{graphicx}
\usepackage{comment}
\usepackage{subfigure}
\usepackage[]{algorithm}
\usepackage[]{algorithmic}
\usepackage{tikz}
\usetikzlibrary{arrows,automata}
\usepackage{multirow}
\usepackage[colorinlistoftodos]{todonotes}
\usepackage[colorlinks=true, allcolors=blue]{hyperref} 
\usepackage{cite}
\usepackage{enumerate}
\usepackage[numbers]{natbib}

\newtheorem{definition}{Definition}

\newif\ifsubmit
\submitfalse
\ifsubmit
\newcommand{\stnote}[1]{} 
\newcommand{\ysnote}[1]{} 
\newcommand{\roma}[1]{} 
\newcommand{\ellie}[1]{} 
\newcommand{\tnnote}[1]{} 
\newcommand{\ngnote}[1]{} 
\else
\newcommand{\stnote}[1]{\textcolor{blue}{\textbf{ST: #1}}} 
\newcommand{\ysnote}[1]{\textcolor{violet}{\textbf{YO: #1}}} 
\newcommand{\roma}[1]{\textcolor{purple}{\textbf{RP: #1}}} 
\newcommand{\ellie}[1]{\textcolor{green}{\textbf{EP: #1}}} 
\newcommand{\tnnote}[1]{\textcolor{teal}{\textbf{TN: #1}}} 
\newcommand{\ngnote}[1]{\textcolor{olive}{\textbf{NG: #1}}} 
\fi

\def\U{\mathcal{U}}
\def\F{\mathcal{F}}
\def\G{\mathcal{G}}
\def\B{\mathcal{B}}
\def\M{\mathcal{M}}
\def\L{\mathcal{L}}

\def\P{\mathcal{P}}

\title{Planning with State Abstractions for Non-Markovian Task Specifications}

\author{
Yoonseon Oh, Roma Patel, Thao Nguyen, Baichuan Huang, Ellie Pavlick, and Stefanie Tellex \\
\thanks{The authors are with the Brown University Department of Computer Science, 115 Waterman Street, Providence, RI 02912. Email: \{yoonseon\_oh, romapatel, thao\_nguyen3, baichuan\_huang, ellie\_pavlick\}@brown.edu, stefie10@cs.brown.edu}}

\begin{document}
\maketitle

\begin{abstract}
Often times, we specify tasks for a robot using temporal language that can also span different levels of abstraction.
 The example command \textit{``go to the kitchen before going to the second floor''} contains spatial abstraction, given that ``floor'' consists of individual rooms that can also be referred to in isolation (``kitchen", for example). There is also a temporal ordering of events, defined by the word ``before".
Previous works have used Linear Temporal Logic (LTL) to interpret temporal language (such as ``before"), and Abstract Markov Decision Processes (AMDPs) to interpret hierarchical abstractions (such as ``kitchen" and ``second floor"), separately. 
To handle both types of commands at once, we introduce the Abstract Product Markov Decision Process (AP-MDP), a novel approach capable of representing non-Markovian reward functions at different levels of abstractions. The AP-MDP framework translates LTL into its corresponding automata, creates a product Markov Decision Process (MDP) of the LTL specification and the environment MDP, and decomposes the problem into subproblems to enable efficient planning with abstractions. 
AP-MDP performs faster than a non-hierarchical method of solving LTL problems in over $95 \%$ of tasks, and this number only increases as the size of the environment domain increases.
We also present a neural sequence-to-sequence model trained to translate language commands into LTL expression, and a new corpus of non-Markovian language commands spanning different levels of abstraction.
We test our framework with the collected language commands on a drone, demonstrating that our approach enables a robot to efficiently solve temporal commands at different levels of abstraction.
\end{abstract}

\section{Introduction}
In an ideal human-robot interaction scenario, humans would give robots tasks in the form of natural language utterances and gestures.
The variation in language used allows for specifying tasks at varying levels of spatial abstractions, while specifying temporal constraints.  
Meaning can be conveyed with language at different levels of spatial abstraction, in
terms of high-level goals (such as \textit{``fly to the end of the first floor"}), lower-level specifications (such as \textit{``fly east, go south, go south and fly east again"}), or mixed-level (such as \textit{``go to the yellow room and the second floor"}). Language can also express explicit constraints on the path taken to reach the goal (for example, \textit{``fly to the red room first, without going through the green room."}). The former category of commands requires an agent to fluidly move within an abstraction hierarchy (that is, knowing that a \textit{floor} is at a higher level than individual rooms and directions), while the latter command restricts the space of possible paths that can be taken and sometimes induces temporal constraints on the order in which goals can be visited. It is crucial for robot systems to portray an adequate understanding of such commands, coupled with the ability to efficiently execute the underlying task.

\begin{figure}
\centering
\includegraphics[width=0.95\columnwidth]{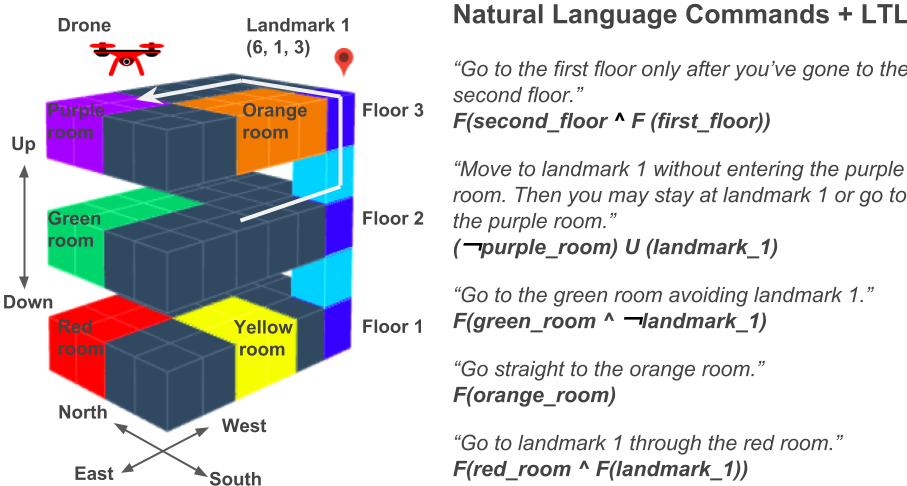}
\caption{Our environment is a gridworld with three floors, each consisting of rooms that consist of grid cells. The white arrow shows an example path the drone can take in the environment. We also include sample natural language commands (and their LTL formulae) that the drone successfully executed.
}
\label{fig:env1}
\end{figure}

Given an environment, a goal condition and constraints, robots can use planning to reach goal conditions while satisfying constraints. Existing approaches interpret language by mapping to a reward function in a Markov Decision Process (MDP) \citep{macglashan2015grounding}. However, these models very quickly become intractable as the state space of the world grows larger \cite{aMDP, konidaris2016}.
Planning with \textit{abstractions} in a hierarchical structure \cite{aMDP, konidaris2018,konidaris2016,sutton1999between}, either by using an Abstract Markov Decision Process (AMDP) \cite{aMDP} or with options \cite{konidaris2018,konidaris2016,sutton1999between} can allow reduction of the state space. There has been previous work in interpreting natural language task specifications at different levels of spatial abstraction and planning using AMDPs \citep{arumugam2017accurately}. 
Separately, as shown in Fig. \ref{fig:env1}, non-Markovian natural language commands can be mapped to linear temporal logic (LTL) formulae \cite{LTLMop,kress2008translating,lignos2015provably,boteanu2016model} to allow efficient planning with an MDP, given the corresponding LTL task specifications \cite{ding2011mdp,ding2014optimal,fu2014probably,kasenberg2017interpretable, wolff2012robust,Sadigh2014,gopalan2018sequence}. 
Combining the interpretation of language using a hierarchical structure and the mapping of commands to LTL expressions is non-trivial, as the non-Markovian constraints might span different levels of abstraction. 
Plans in a more abstract state space could therefore lead to failure of constraints specified in a less abstract space (that is, plans at a lower level in the abstraction hierarchy).

\begin{figure}
    \centering
    \includegraphics[width=0.95\columnwidth]{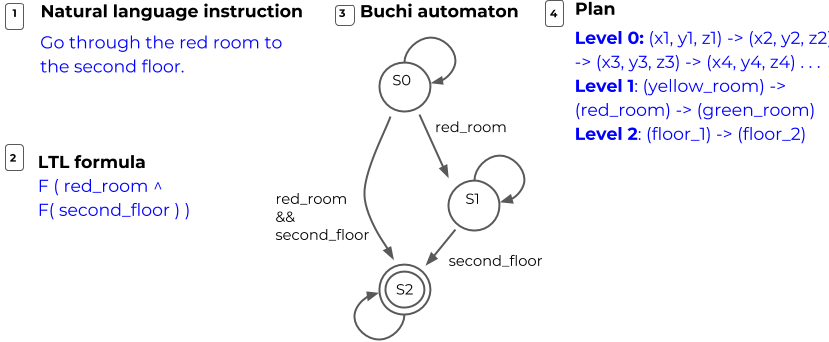}
    \caption{Complete pipeline for the translation of a natural language instruction to an LTL formula, then to a B\"uchi automaton, and to a plan that gives us action sequences to correctly reach the goal location specified by the task.}
    \label{fig:pipeline}
\end{figure}

In this paper, we introduce the {\it Abstract Product MDP (AP-MDP)} framework to combine the benefits of LTL and AMDP, thus enabling a robot to interpret non-Markovian commands at different levels of abstraction. There is previous work in planning for LTL tasks using options \cite{liu2018compositional}. However, the AMDP approach suits our task better, as its hierarchical structure closely resembles the hierarchies formed by humans when planning to solve complex tasks that can be decomposed into subtasks \cite{aMDP}.
In our approach, task specifications are first given as natural language utterances that are then translated into LTL expressions by a supervised neural sequence-to-sequence model.
This LTL expression $\phi$ is converted into a finite state representation that accepts infinite inputs, or a B\"uchi automaton \cite{buchi1990decision}.  This representation allows us to decompose  the problem into sub-problems (organized around sub-parts of the input LTL expression). Edges of the B\"uchi automaton consist of atomic propositions in expression $\phi$ and a sub-problem induces a state transition of the automaton. To further deal with different levels of abstraction, if atomic propositions in the same edge are from different levels, we solve the sub-problem using the lower level AMDP. The robot must then forgo the computational benefits of the AMDP to guarantee that the policy satisfies all the constraints present in the LTL expression.  This entire pipeline (shown in Fig. \ref{fig:pipeline}) therefore fluidly allows complex task specifications with non-Markovian constraints to be specified using natural language and solved at different levels of the goal hierarchy.  

We evaluate our approach by reporting the performance of AP-MDP in simulation and on a drone platform. We also present a new corpus of non-Markovian natural language commands at different levels of abstraction, a neural sequence-to-sequence model that translates human-uttered natural language commands to their corresponding LTL counterparts, and demonstrates the solving of complex natural language task specifications using AP-MDP on a drone.

\section{Related Work}
LTL has been used to model agent behavior in planning problems with non-Markovian task specifications. Consider a task that requires an agent to visit regions of interest in a specific order (for example, \textit{``visit the red room first, then the blue room, and the green room last"}). These kinds of expressions have intrinsic temporal information that must be taken into account when determining the kind of path that has to be taken to achieve the goal. LTL allows us to formally describe these kinds of task specifications as logical functions, thus allowing robots to then execute these behaviors.

When the goal for a task is defined as an LTL expression, previous works have often formulated the problem as a product of an MDP and an automaton of the LTL formula \cite{ding2011mdp, ding2014optimal,fu2014probably,wolff2012robust,Sadigh2014}.
Some previous works model dynamic systems of agents as MDPs and developed methods to generate a control policy that satisfies LTL constraints \cite{ding2014optimal, ding2011mdp}. The LTL formula is converted into a {\it Deterministic Rabin Automaton} (DRA), 
and the dynamic system is formulated as a product of a DRA and MDP. The goal is then to search for a policy that satisfies the acceptance condition. Along the same lines, Kasenberg and Scheutz \cite{kasenberg2017interpretable} show that the reverse is also true, that is, the product of a DRA and an MDP can be considered to infer an LTL specification from demonstrations. However, this approach does not scale well for large MDPs. 

Decision making with an MDP often becomes intractable as the size of the state space increases. In order to overcome intractability, hierarchical frameworks \cite{konidaris2016,konidaris2018, nips2016Tejas, aMDP} are commonly used. The options framework \cite{konidaris2016,konidaris2018}, for example, models temporally abstract macro-actions as options that can be adopted to build abstraction hierarchies.
Similarly, AMDPs \cite{aMDP} can be used for abstraction by decomposing tasks into series of subtasks, thus allowing planning to take place more efficiently. However, these methods do not address the problem of solving LTL specifications with abstractions. 

Hierarchical frameworks are powerful when an agent is faced with the task of planning a sequence of actions for complex LTL tasks. Several works \cite{Fainekos2009heirarchical,McMahon2014sampling,kyunghoon,CDC2017oh} propose incorporating both the robot dynamics and the given LTL constraints in a continuous space. 
A continuous state space can be abstracted into a discrete state space and a continuous path is derived by sampling guided by the high-level discrete plan \cite{McMahon2014sampling,kyunghoon,CDC2017oh}. 
Other works have focused on grounding natural language to LTL expressions \cite{gopalan2018sequence,lignos2015provably,boteanu2016model} to further allow a robot to make use of these LTL specifications. Previous work in hierarchical planning using options can accelerate planning for LTL tasks \cite{liu2018compositional}. However, the AMDP framework  \cite{aMDP} is better suited for our task than options, by virtue of encoding a goal hierarchy rather than learning a policy over goals.

To the best of our knowledge, this work is the first to propose a hierarchical framework for planning for LTL tasks using the structure of an AMDP. 
An AMDP provides abstract states, actions, and transition dynamics in multiple layers above a base-level MDP, thus decomposing problems into subtasks with local rewards and local transition functions for policy generation. Moreover, as shown in our robot demonstrations, we start from human input given in the form of speech that is then converted to text. This textual input of the natural language command is translated to its LTL representation, and atomic propositions are directly mapped into propositions in each layer of a multi-level AMDP. We can then plan at levels higher than the lowest level whenever possible, and find a policy in a more efficient way than previous approaches.

\section{Problem Formulation}
We consider a planning problem for a robot, when the task that the robot is required to interpret and solve is given through a natural language command. Our environment is a 3D grid world consisting of three floors as shown in Fig. \ref{fig:env1}. Each floor is composed of colored rooms, a room is composed of a set of grid cells, a landmark (such as a charging station) indicates a cell at position \textit{(x,y,z)}.
Landmarks (or cells) are therefore the lowest level of abstraction, rooms are abstract expressions of landmarks, and floors are abstract expressions of rooms and form the highest level of abstraction. A natural language command (such as \textit{``first go to the red room through landmark 1 and then go to the blue room."}) is given to the robot by virtue of observable visual elements in our abstraction hierarchy (landmarks, rooms, and floors). This natural language utterance is grounded to its LTL counterpart ($\F$ (\textit{landmark\_1} $\wedge$ $\F$ (\textit{red\_room} $\wedge$ $\F$ (\textit{blue\_room})))) which forms the task specification. The agent is required to accomplish the task by correctly finding a path to the correct location and following the determined path by executing a sequence of actions from the action set (\textit{north, south, east, west, up, down}).

We formulate this problem as an MDP that gets a high reward when the task is accomplished. Crucially, we make use of abstractions over the MDP state space for more efficient planning in large environments, and for the robot to efficiently find policies for commands at different levels of abstraction.  
Consider the example task above of \textit{``first go to the red room through landmark 1 and then go to the blue room."} This is an expression that spans different levels in the abstraction hierarchy (that is, rooms and landmarks) and can be translated into its equivalent LTL formula $\phi$ over atomic proposition sets $AP^L$ for each level $L$ in the hierarchy. For example, \textit{``landmark 1"} occupies one grid cell in the environment and corresponds to an atomic proposition (denoted by $\alpha^0_0$) in $AP^0$ and \textit{``red room"} and \textit{``blue room"} correspond to atomic propositions (denoted by $\alpha_0^1$ and $\alpha_1^1$, respectively) in $AP^1$.
The expression can be translated into 
$\phi=\F(\alpha_0^0 \wedge \F(\alpha_0^1 \wedge \F \alpha_1^1))$ using the LTL operator $\F$ or ``finally", converted to a B\"uchi automaton \cite{buchi1990decision, LTLMop}, and then an AMDP \cite{aMDP} to decompose the problem into a series of smaller, and hence easier to solve, subproblems.
Section \ref{sec:prelim} defines LTL and the variants of MDPs that our model relies on, while section \ref{sec:ap-mdp} goes over how they are composed together to produce a more efficient solution, while describing the end-to-end pipeline with the natural language grounding components. 



\section{Preliminaries}\label{sec:prelim}
This section defines the components used in our formulation and how they are transformed into one another to form state abstractions for complex, non-Markovian task specifications uttered by humans through natural language. We briefly introduce LTL and its syntax, explain the transformation of an LTL expression to a B\"uchi automaton and further to an MDP. 


\subsection{Linear temporal logic}
Temporal logic was first introduced as a formalism for clarifying issues of time and defining the semantics of temporal expressions. LTL is a temporal logic whose syntax contains path formulae --- the logical expression describes a specification that can be validated over a trajectory of any robot (discrete) system. LTL has the following grammatical syntax:
$\phi::= \pi~|~\neg \phi~|~\phi \wedge \varphi~|~\phi \vee \varphi~|~\G \phi~|~\F \phi~|~ \phi~\U \varphi $, 
where $\phi$ is the task specification or path formula, $\phi$ and $\varphi$ are both LTL formulae, $\pi \in \Pi$ is an atomic proposition, $\F$ denotes ``finally", $\G$ denotes ``globally" or ``always", $\U$ denotes ``until", and $\neg, \wedge, \vee$ denote logical ``negation", ``and" and ``or". 


\subsection{Linear temporal logic to B\"uchi automaton}
An LTL formula intuitively expresses properties over trajectories or traces (a sequence of sets of atomic propositions) in the environment. This can be translated into an equivalent B\"uchi automaton \cite{buchi1990decision} --- a deterministic automaton, that differs from the general notion of automata in that it accepts infinite traces represented by the input LTL formula. This handling of infinite traces is specifically necessary in cases of complex non-Markovian task specifications that can map to potentially unbounded action sequences.

\begin{definition}
(B\"uchi automaton): A deterministic B\"uchi automaton (DBA) is a tuple $\B=(Q,\Sigma,\delta,q_0,{\bf F})$ where $Q$ is a finite set of states, $\Sigma$ is the input alphabet, $\delta: Q \times \Sigma \rightarrow Q$ is the transition function, $q_0\in Q$ is the initial state, and ${\bf F}$ is the acceptance condition.
\end{definition}

For the LTL formula $\phi$, the input alphabet of the automaton $\B$ is $\Sigma=2^{AP}$. A word $w$ over an alphabet can be any infinite sequence of atomic propositions, and the \textit{run} of the automaton on $w=a_0a_1\cdots$ with $a_i\in \Sigma$ is a sequence of states $\rho=q_0q_1\cdots$ for $q_i\in Q$, where $q_0$ is an initial state and $q_{i+1}=\delta(q_i,a_i)$. A word is accepted by the automaton iff its run $r$ satisfies the relationship $\lim(r)\cap {\bf F}\not= \emptyset$, that is, the language $L(\B)$ is non-empty if at least one final state is reached.


\subsection{Labeled Markov Decision Processes}

In order to combine an MDP with the LTL formula to make an expanded MDP, we need to annotate each state with propositions so that we can evaluate the LTL expression.
A labeled MDP \cite{fu2014probably} is essentially an MDP where transitions are annotated with labels. These labels are provided by a labeling function that maps states to valid propositions for each state.

\begin{definition}
(Labeled MDP): A labeled MDP is a tuple $\M = (S,A,T,s_0,AP,L,R)$, where $S$ and $A$ are finite state and action sets, $T:S\times A\times S\rightarrow[0,1]$ is a transition probability function, $s_0\in S$ is the initial state, $AP$ is a set of atomic propositions, $L:S\rightarrow 2^{AP}$ is a labeling function which maps a state $s\in S$ into a set of atomic propositions valid at state $s$, and $R: S\rightarrow \mathbb{R}$ is a reward function.
\end{definition}
\subsection{Product Markov Decision Processes}\label{sec:productMDP}

We now need to combine the labeled MDP $\M$ with the LTL expression in order to make an expanded MDP which keeps track of the relevant parts of the LTL state.
A product automaton is one that derives from the product of the finite transition system of $\M$ and the automaton $\B$ that represents the LTL specification. Labeled MDPs have previously been used for planning over an MDP to satisfy an LTL formula \cite{wolff2012robust,Sadigh2014}, where the states of $\M$ and $\B$ encode the desired LTL specification. We can therefore design a state based reward function that relies on acceptance conditions of $\B$.
\begin{definition}
(Product MDP): Given a deterministic B\"uchi automaton $\B=(Q,\Sigma,\delta,q_0,{\bf F})$ and a labeled finite MDP $\M = (S,A,T,s_0,AP,L,R)$
, with $s\in S$ and $q\in Q$, the product MDP (P-MDP) for the state $(s,q)$ is given by $\M_p = (S_p, A, T_p, s_{0p},Q, L_p)$ where:

\begin{enumerate}[(a)]
\item $S_p = S\times Q$ is a product state,
\item $T_p((s,q), a, (s',q'))=~\left\{ \begin{array}{l}T(s,a,s'),\text{ if }q'=\delta(q,L(s'))\\0,\text{~~~~~~~~~~~otherwise,} \end{array}\right.$
\item $s_{0p} = (s_0,q)$ such that $q = \delta(q_0,L(s_0))$, 
\item $L_p((s,q))={q}$,
\end{enumerate}
\end{definition}

\subsection{Abstract Markov Decision Processes}
An Abstract Markov Decision Process \cite{aMDP} (AMDP) hierarchy decomposes large planning problems into a series of subproblems with local reward and transition functions using state and action abstraction. 

\begin{definition}
(Abstract MDP): An AMDP is a 6-tuple $\tilde{\M} = (\tilde{S},\tilde{A},\tilde{T},\tilde{R},\tilde{E},F)$. These are the usual MDP components, with the addition of $F : S \rightarrow \tilde{S}$, a state projection function to map states from the original environment MDP into the AMDP abstract state space $\tilde{S}$. Actions in the action set $\tilde{A}$ of the AMDP are either primitive actions, or are associated with subgoals to solve in the environment MDP. The transition function $\tilde{T}$ captures the dynamics of the effects of changes in the AMDP state space once subgoals are completed. $\tilde{R}$ is the reward function. $\tilde{E} \subset \tilde{S}$ is the set of terminal states.
\end{definition}

\section{Technical Approach}\label{sec:ap-mdp}
At a high level, we use a neural sequence-to-sequence model to convert an English command to the corresponding LTL expression, which is then translated to a B\"uchi automaton and then levels of the component AMDP to enable the robot to infer a policy based on the expression. We run a simulation that shows the produced action sequence, executable by a drone in a 3D environment.


\subsection{Abstract Labeled Markov Decision Processes}
We propose {\it Abstract Labeled MDPs (AL-MDPs)} that decomposes an MDP $\M$ into multiple abstract labeled MDPs which are based on abstract states, actions, and transitions in multiple layers.
The labeled MDPs in the lowest level, the $i$th level, and the highest level are denoted by $\hat\M^0$, $\hat\M^i$, and $\hat\M^L$, respectively. The abstract labeled MDP $\hat\M^i$ is defined below: 

\begin{definition} (Abstract Labeled MDP):~$\hat\M^i=(\hat{S}^i,\hat{A}^i,\hat{T}^i,\hat{s}^i_0,AP,\L^i,R^i)$, 
where $\hat{S}^i,~\hat{A}^i,~\hat{T}^i$ and $R^i$ are a set of states, a set of actions, a transition function, and a reward function, respectively.
States in $\hat\M^i$ correspond to a combination of atomic propositions in $AP$ by the labeling functions $\L^i: \hat{S}^i\rightarrow 2^{AP}$.
The set of atomic propositions $AP$ is a union of $L$ disjoint sets $AP^i$s, where $AP^i=\{\alpha_0^i,\cdots,\alpha_n^i\}$ (that is, $AP = \cup_{i=1}^L AP^i$).
The proposition $\alpha\in AP$ belongs to $AP^i$, where $i$ is the largest value which satisfies that there exists a state $s\in\hat{S}^i$ which can determine the truth value of $\alpha$.
\end{definition}

\subsection{Abstract Product Markov Decision Processes}
We propose {\it Abstract Product MDPs} (AP-MDPs) which combine AL-MDPs and DBAs to solve ordinary product MDPs efficiently. We furthermore show how our approach handles a combination of atomic propositions in multiple levels. For example, if some of the atomic propositions are defined at level $0$, we cannot guarantee that a plan derived at level $1$ or level $2$ will satisfy level $0$ constraints. This would require working at the lowest level of atomic propositions, thus losing the computational benefit of abstraction and a reduced state space. In all previous hierarchical approaches in this area, when atomic propositions of different levels exist together, the product MDP must be solved at the lowest level (level $0$ in this case) to guarantee the satisfaction of the transition constraint that directly affects it. This therefore does not afford the computational benefit of planning at higher levels using AL-MDPs. Our approach, however, employs different depths of AL-MDPs by decomposing the product MDP into subproblems to benefit from the hierarchical structure when the LTL task includes atomic propositions at the lowest level.

AP-MDPs combine the automaton $\B$ of the LTL task specification with AL-MDPs. This involves taking an LTL formula in the form of an automaton, converting it to a labeled MDP and decomposing this MDP into several subproblems, each of which are individually solved at the required level of abstraction.
 We use a running example, as shown in section \ref{subsec:example} to highlight the process of how decomposed subproblems are solved for the task specification in question. Section \ref{subsec:decompose} then explains how any problem can be decomposed into component subproblems and section \ref{subsec:alg} presents the pseudocode for the algorithm for this process. The language grounding component of the system is discussed in \ref{subsec:langgrounding} and finally, \ref{subsec:planning} describes the functioning of the end-to-end system.

\subsection{Example Problem}\label{subsec:example}
\begin{figure}[t]
\centering
\includegraphics[height=.4\columnwidth]{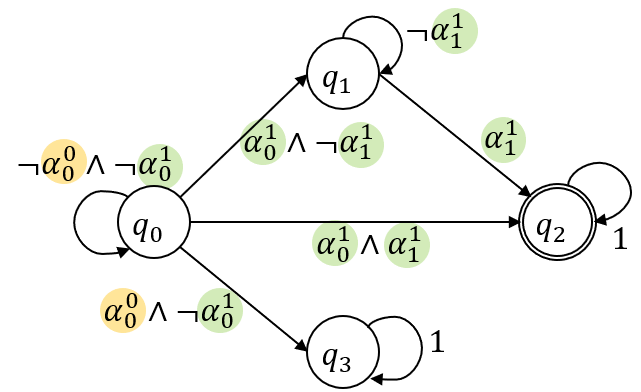}
\caption{Deterministic B\"uchi automaton. Atomic propositions in yellow circles correspond to those in level $0$ and atomic propositions in green circles correspond to those in level $1$. The transitions of the automaton refer to constraints over the propositions that are satisfied on taking that path.}
\label{fig:DBA_example}
\end{figure}

Consider the example in Fig. \ref{fig:DBA_example}. This figure shows the DBA for the LTL task specification $\phi = \F(\alpha_0^0\wedge\F(\alpha_0^1\wedge\F \alpha_1^1))$ and we can see that the atomic proposition $\alpha_0^0$ is in level $0$ of the abstraction hierarchy, while $\alpha_0^1$ and $\alpha_1^1$ are in level $1$. To deal with these different levels in the abstraction hierarchy, we decompose the entire problem into different subproblems.
The first subproblem $\hat{\M}_0$ is defined by a tuple  $\hat{\M}_0=(\hat{S}^0, \hat{A}^0, \hat{T}^0, \hat{s}^0_0, AP, \L^0, R^0)$ and here the agent wants to go to $q_1$ while not visiting other states in the DBA. The condition to reach the desired state, $f(q_0,q_1,s,s')=true$ is its {\it goal condition} and the condition to stay in the current state, $f(q_0,q_0,s,s')=true$ is its {\it stay condition}, where $s$ and $s'$ are the current state and the next state, respectively.
The function $f$ returns $true$ or $false$ depending on whether the logical expression on the edge is satisfied by the state.
The reward function ensures that the agent gets a large positive reward if the goal condition is satisfied and gets a large negative reward if the stay condition is violated and the goal condition is not satisfied. In all other cases, it gets a small negative reward as the time taken increases.  Since this subproblem contains atomic propositions at level $0$, we can solve it at level $0$, that is, the lowest level of atomic propositions.

We now consider the latter part of the decomposition, that is, the second subproblem $\hat{\M}_1$. This has atomic propositions related to level $1$, therefore $\hat{\M}_1$ can be formulated at a higher level of abstraction, that is, level $1$
($\hat{\M}_1=(\hat{S}^1,~\hat{A}^1,~\hat{T}^1,~\hat{s}^1_0,~AP,~\L^1,~R^1)$), allowing for more efficient planning over a smaller state space. In this way, all subproblems $\hat{\M}_i$ can be solved at the desired level to allow for full use of the benefits of abstraction where possible.

\subsection{Subproblem Decomposition}\label{subsec:decompose}
In general, there are $n_\rho$ paths in a B\"uchi automaton from the initial state to the accepting state. AP-MDPs decompose the problem into $n_\rho$ subproblems each denoted by $\P_{\rho_i}$, which accomplish the LTL task while following the path $\rho_i$.

Each problem $\P_{\rho_i}$ can be  decomposed into $n_i$ subproblems, each formulated by an AL-MDP. Each $\P_{\rho_i}^j$s aims to change the DBA state of the agent from $\hat{q}_{j}^i$ to $\hat{q}_{j+1}^i$ and the {\it goal condition} and the {\it stay condition} of $\P_{\rho_i}^j$ are $f(\hat{q}_{j}^i,\hat{q}_{j+1}^i,s,s')=true$ and $f(\hat{q}_{j}^i,\hat{q}_{j}^i,s,s')=true$, respectively.  
The reward function for the AL-MDP is defined by:
\begin{eqnarray}
R_j = \left\{\begin{array}{ll}
\gamma_{goal}, & \text{if } f(\hat{q}_j^i,\hat{q}_{j+1}^i,s,s') = true,\\
\gamma_{stay}, & \text{else if } f(\hat{q}_j^i,\hat{q}_{j}^i,s,s') = false, \\
\gamma, & \text{otherwise,} 
\end{array}\right.
\end{eqnarray}
where $\gamma_{goal} \gg 1$, $\gamma_{stay} \ll 0$, and $\gamma$ is a small negative value. 

In this way, AP-MDPs can consist of ($\sum_{i=1}^{n_\rho}{n_i}$) AL-MDPs. When we denote the plan for $\P_{\rho_i}$ as $(s_{seq}, a_{seq})^{\rho_i}$,
the plan for the LTL task is the shortest sequence $(s_{seq}, a_{seq})^*$, where $s_{seq}$ is the state sequence and $a_{seq}$ the is action sequence.

\subsection{Algorithm}\label{subsec:alg}
\begin{algorithm}[htb]
\begin{algorithmic}[1]
\STATE LTL task $\phi$ and $s_0$ are given
\STATE Initialize the optimal plan, $(s_{seq}, a_{seq})^*$.
\STATE Initialize the length of the optimal plan, $l^*$.
\STATE $A \leftarrow LTL2DBA(\phi)$ \label{line:spot2} 
\STATE $A.RemoveContradiction()$\label{line:simplify}
\STATE $Paths = A.FindPaths()$ \label{line:findpaths}
\FOR{$\rho_i \in Paths$} 
	\STATE Initialize $s_0$    
	\STATE Initialize the plan $(s_{seq},~a_{seq})^{\rho_i}$ 
	\FOR{$j$ in $\{0,\cdots,n_i-1\}$}
 	   \STATE {\it goal condition} $\leftarrow$ $f(\hat{q}_j^i,\hat{q}_{j+1}^i,s,s') = true$ \label{line:goal}
       \STATE {\it stay condition}$\leftarrow$ $f(\hat{q}_j^i,\hat{q}_j^i,s,s') = true$
    	\STATE $\ell_j \leftarrow$ the lowest level of atomic propositions in {\it goal} and {\it stay conditions}.  
        \STATE $\hat{\M}_j\leftarrow(\hat{S}^{\ell_j},~\hat{A}^{\ell_j},~\hat{T}^{\ell_j},~\hat{s}^{\ell_j}_0,~AP,~\L^{\ell_j},~R^{\ell_j}).$ \label{line:Mj}
        \STATE $\pi \leftarrow$ $\hat{\M}_j.Solve()$ \label{line:solve}
        \STATE $ss, aa\leftarrow \hat{\M}_j.Plan(\pi, ~s_0)$ \label{line:plan}
        \STATE $(s_{seq},~a_{seq})^{\rho_i} \leftarrow (s_{seq},~a_{seq})^{\rho_i}~ \cup  (ss,aa)$
        \STATE $s_0 \leftarrow s_{seq}(end)$
    \ENDFOR
\IF{$length(s_{seq}) < l^*$} \label{line:optimal_begin}
\STATE $(s_{seq},a_{seq})^* \leftarrow (s_{seq},a_{seq})^{\rho_i}$
\ENDIF \label{line:optimal_end}
\ENDFOR
\end{algorithmic}
\caption{Solve AP-MDPs}\label{alg:AP-MDPs}
\end{algorithm}
The entire algorithm is presented as pseudocode in Algorithm \ref{alg:AP-MDPs}.
The input task is specified as an LTL expression composed of atomic propositions in the environment and the logical operators defined previously. We translate the LTL formula into a DBA using an existing package called Spot2 (line \ref{line:spot2}) \cite{spot2}.
Note that the DBA may contain infeasible edges because the translator does not consider the real environment (for example, if the \textit{red\_room} does not exist on the \textit{first\_floor} in a particular gridworld, $\texttt{red\_room}\wedge\texttt{floor\_1}$ cannot be $true$). We handle this by eliminating edges which have contradictions consisting of a logical incompatibility between two or more propositions (line \ref{line:simplify}), based on specifications of the environment in question.
We check the contradiction by looking at the truth table of the formula.
We then find all possible paths from the initial state to the accepting state in line \ref{line:findpaths}.
The AL-MDPs \textit{goal} and \textit{stay} conditions are defined through lines \ref{line:goal} to \ref{line:Mj}, and we then obtain the optimal policy and plan of AL-MDPs with a solver of the AMDP (lines \ref{line:solve}-\ref{line:plan}).
We then select the best plan which has the minimum number of actions (lines \ref{line:optimal_begin}-\ref{line:optimal_end}).

\begin{figure*}[htb]
    \centering
    {\includegraphics[width=1.75\columnwidth]{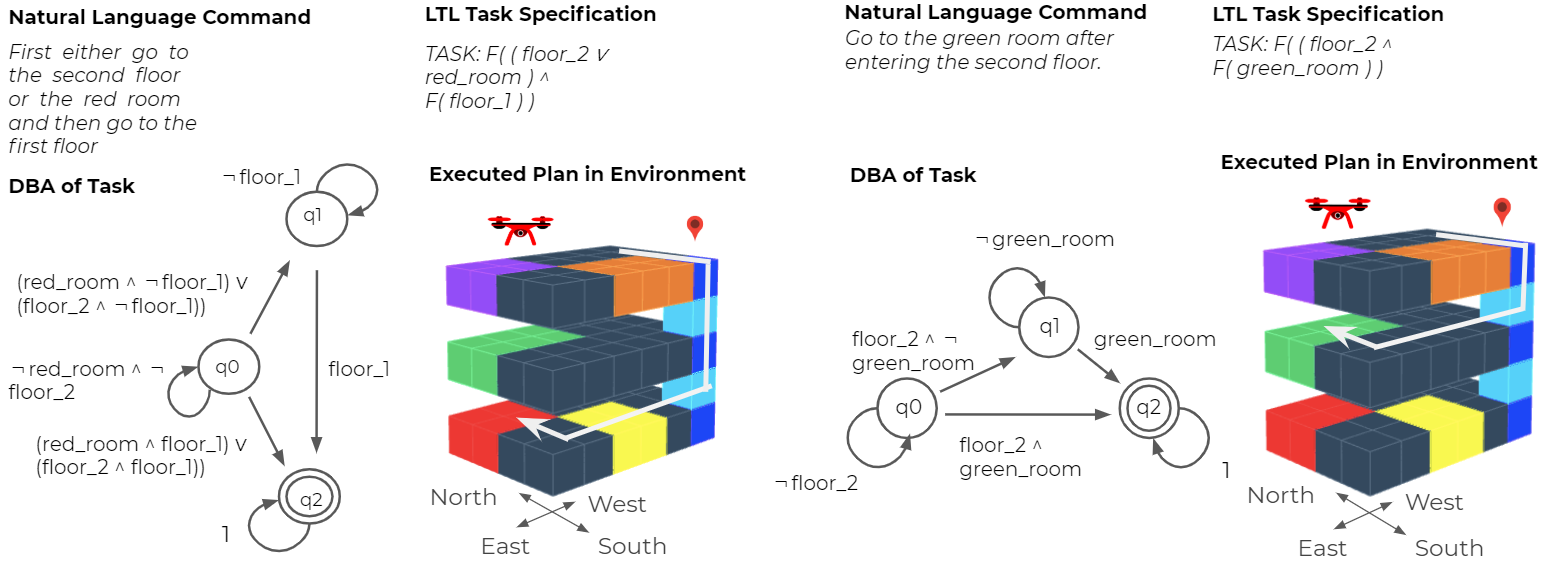}}
    \caption{Examples, left and right, tested in simulation. In each example, a natural language instruction is converted to an LTL expression, then to a corresponding AP-MDP to find a policy. An agent then executes the policy in the specified environment to reach the correct goal state through the desired path.}
    \label{fig:task34}
\end{figure*}


\subsection{Grounding language to LTL formulae}
\label{subsec:langgrounding}
We train a neural sequence-to-sequence model to translate natural language commands to LTL expressions. We discuss our language corpus and the model architecture below.
\subsubsection{Corpus}
We use Amazon Mechanical Turk (AMT) to collect non-Markovian natural language commands that also refer to elements in the environment at different levels of abstraction\footnote{The corpus can be found at \url{https://github.com/h2r/ltl-amdp}}. 
AMT workers were shown images representing correct and incorrect ways for the robot to complete a task, and asked to give commands that accurately capture the robot's correct behavior. $810$ natural language commands were collected from $120$ AMT workers for $27$ LTL formulae. We augment these $810$ commands to obtain $6185$ commands for $343$ LTL expressions. Augmentation is done by mapping one training sample (for example, \textit{``go to the red room''} accompanied by $\F(red\_room)$) to similar commands and corresponding LTL expressions for every other possible goal locations. We held aside $20\%$ of the data as the test set to evaluate model performance and trained on all remaining data and perform 5 fold cross-validation in this manner.  
\subsubsection{Sequence-to-sequence model}
As in Gopalan et al. \cite{gopalan2018sequence}, we use a neural sequence-to-sequence model composed of a recurrent neural network (RNN) encoder and decoder to translate each natural language instruction to an LTL formula. It is implemented in PyTorch \cite{paszke2017automatic} and trained for $10$ epochs over our corpus, with a learning rate of $0.001$ using the Adam optimizer \cite{kingma2014adam}. We used a dropout of $0.8$ as a regularizer~\cite{srivastava2014dropout}.

\subsection{Planning for an LTL task}
\label{subsec:planning}
Once a natural language command is translated into an LTL formula, it is then converted into a B\"uchi automaton with multiple paths from the initial state to the accepting state. Each path is represented and solved with an AL-MDP.
\section{Experiments}
In this section, we show that our method efficiently generates plans for complex LTL tasks. 
We evaluate efficiency with the number of backups and the computation time over $100$ tasks. 
We successfully applied the proposed method on a drone.

\subsection{Environment Setup}
For simulations, we consider two 3D grid worlds ($\mathcal{E}_1$ and $\mathcal{E}_2$) of size $6\times4\times3$ and $30\times20\times6$, respectively. The smaller world $\mathcal{E}_1$ has three floors, each comprised of six rooms, each the size of $2 \times 2$ grid cells. The larger world $\mathcal{E}_2$ has six floors, each comprised of six rooms of size $10 \times 10$. The visually observable elements (grid cells, rooms and floors) form the atomic propositions of the LTL task specifications. Importantly, these elements span different levels of abstraction: landmarks (grid cells) are at level $0$, rooms are at level $1$, and floors are at level $2$. While our simulation environments consist of at least three floors, our robot demonstration is performed in a gridworld with only two floors for compatibility with the maximum height our PiDrone can reach.

\subsection{Examples in simulation}\label{sec:sim-example}

We consider the tasks below to demonstrate example simulations of our proposed method. We show the language command with the corresponding LTL task specification, the automaton of the LTL expression, and the path found by our proposed approach for each example. This highlights how our method solves a given task while satisfying the constraints of the task. The tasks in question exhibit the complex constraints with non-Markovian nature and varying levels of abstraction as outlined above. They contain propositions at different levels in the abstraction hierarchy, and contain temporal order constraints by specifying certain subtasks that should be performed before others. The two tasks are:
\begin{enumerate}
\item $\phi_1 \!\!= \!\!\F\left((\texttt{floor\_2}~\vee~\texttt{red\_room})~\wedge~\F (\texttt{floor\_1})\right)$ \\ \textit{(``First either go to the second floor or the red room, and then go to the first floor")} 
\item $\phi_2 = \F (\texttt{floor\_2}~\wedge~\F(\texttt{green\_room}))$ \\ 
\textit{(``Go to the green room after entering the second floor")} 
\end{enumerate}

The execution of both tasks is shown in Fig.~\ref{fig:task34}. The process to solve task $\phi_1$ for the given LTL task specification is outlined in the left side of the figure. Upon decomposing this task specification as in our proposed method, there are two paths of automaton states. Consider the path $\rho_0=q_0q_2$ corresponding to the AL-MDP $\hat{\M}_0$. This has a goal condition of $((\texttt{red\_room}~\wedge~\texttt{floor\_1}) \vee (\texttt{floor\_2}~\wedge~\texttt{floor\_1}))$ and a stay condition of $(\neg\texttt{floor\_2}~\wedge~\neg\texttt{red\_room})$. For the path $\rho_1=q_0q_1q_2$, there are two AL-MDPs $\hat{\M}_0$ and $\hat{\M}_1$, where
$\hat{\M}_0$ has a goal condition of $((\texttt{red\_room}~\wedge~\neg\texttt{floor\_1}) \vee (\texttt{floor\_2}~\wedge~\neg\texttt{floor\_1}))$ and a stay condition of $(\neg\texttt{floor\_2}~\wedge~\neg\texttt{red\_room})$, and $\hat{\M}_1$ has a goal condition of $(\texttt{floor\_1})$ and a stay condition of $(\neg\texttt{floor\_1})$.
Since we can satisfy $\phi_1$ with only two actions with $\rho_0$, 
the final solution is a plan for $\rho_0$.

\begin{figure}
\centering
\subfigure[]{\includegraphics[width=.48\columnwidth]{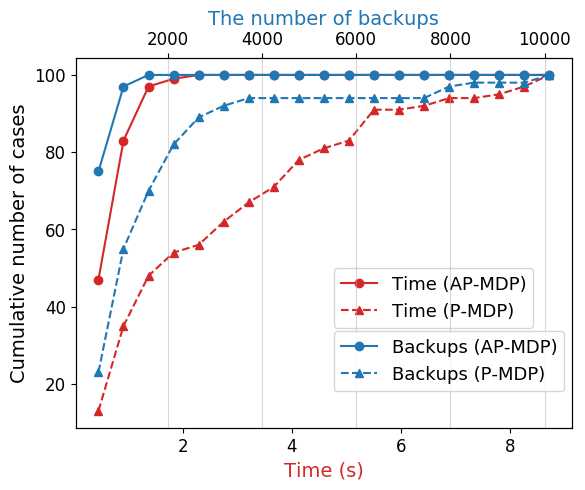}\label{fig:result_small}}
\subfigure[]{\includegraphics[width=.48\columnwidth]{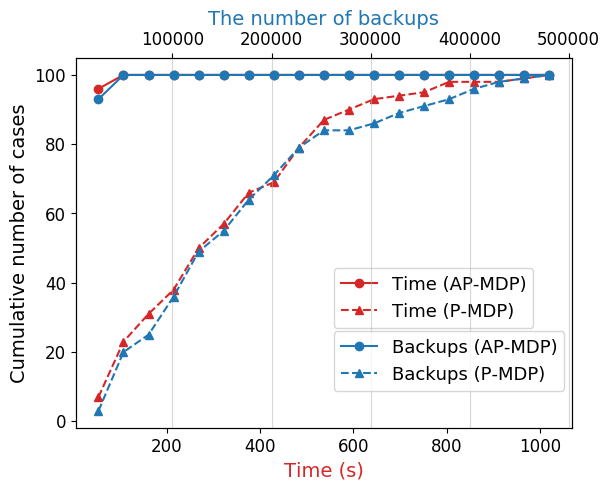}\label{fig:result_large}}
    \caption{Cumulative histograms of computing time and the number of backups of AP-MDP and P-MDP in the environment (a) $\mathcal{E}_1$ and (b) $\mathcal{E}_2$. We execute AP-MDP and P-MDP with $100$ random LTL tasks in two environments, $\mathcal{E}_1$ and $\mathcal{E}_2$. The y-axis shows the cumulative number of cases evaluated.}
    \label{fig:hist}
\end{figure}

For task $\phi_2$, there exists an infeasible path among paths in the automaton. The first AL-MDP in $\rho_0=q_0q_2$ has goal and stay conditions of  $(\texttt{floor\_2}~\wedge~\texttt{green\_room})$ and $(\neg\texttt{floor\_2})$, respectively. This problem does not have a solution because the green room is on the second floor, and thus our algorithm does not return a plan.
There is, however, a solution for the path $\rho_1=q_0q_1q_2$. The first AL-MDP has a goal condition of $(\texttt{floor\_2}~\wedge~\neg\texttt{green\_room})$ and a stay condition of $(\neg\texttt{floor\_2})$.
The second AL-MDP has a goal condition of $(\texttt{green\_room})$ with a stay condition of $(\neg\texttt{green\_room})$.
The planned path is shown in Fig. \ref{fig:task34}.

\subsection{Efficiency}\label{sec:efficiency}
In this section, we evaluate the efficiency of the proposed algorithm by measuring the computing time and the number of backups of the algorithm. The measured computing time includes pre-processing time like translating the LTL expression to a DBA and searching for a path in the DBA, along with the final planning time. The hierarchical structure allows for more efficient planning when unnecessary backup across multiple levels of the hierarchy is limited. We also evaluate the ability of different models to plan without this unnecessary computation. For each problem, the number of backups depends on the number and size of subproblems.

Since planning for an LTL task can be formulated as the product of an automaton $\B$ and MDP $\M$ as described in section \ref{sec:productMDP}, our baseline algorithm (called P-MDP) is one that solves the product MDP at level $0$ using value iteration. We ran $100$ random tasks in the aforementioned environments ($\mathcal{E}_1$ and $\mathcal{E}_2$).
The example tasks here are LTL specifications randomly sampled from the set $\left\{ \F a,~\F(a \wedge \F b),~\F(a \wedge \F (b \wedge \F c)),~\F a \wedge \F b,~\neg a~\U~b\right\}$~, where $a$, $b$, and $c$ are atomic propositions that can be visually observed in our environment (such as \texttt{landmark\_1, green\_room, first\_floor}). We ensure that atomic propositions are sampled from all possible landmarks, rooms, and floors to get a full variety of commands, and ensure that environment constraints are satisfied. For example, if level $1$ is sampled, we sample the index of rooms among all possible rooms in that level. The lowest level of sampled atomic propositions is denoted by $0$. 

\begin{figure}
\centering
\subfigure[Computing time ratio]{\includegraphics[width=.48\columnwidth]{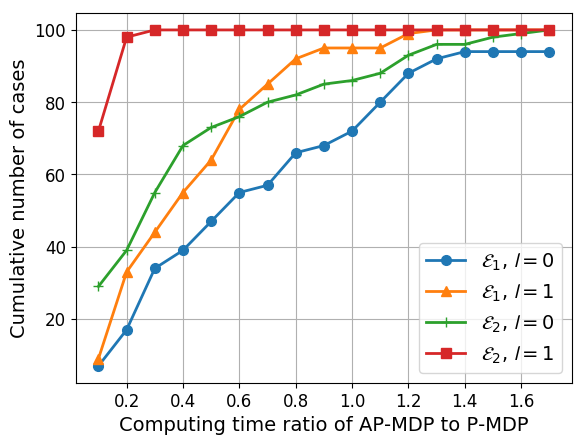}\label{fig:ratio_time}}
\subfigure[The number of backups ratio]{\includegraphics[width=.48\columnwidth]{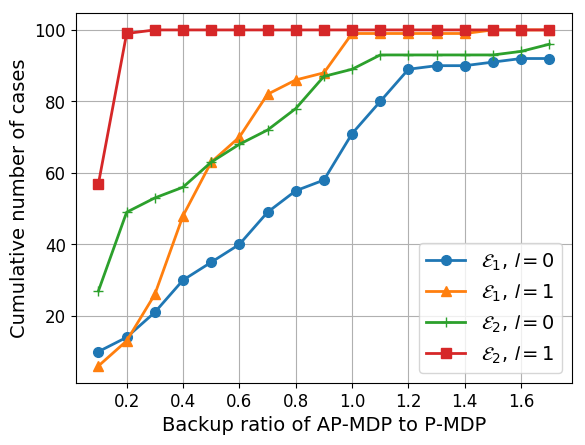}\label{fig:ratio_backup}}
    \caption{Cumulative histograms of (a) computing time ratio (lower is better) and (b) the number of backups ratio (lower is better) of AP-MDP to P-MDP.}\label{fig:ratio}
\end{figure}
 
We display the results as histograms plotted in Fig. \ref{fig:hist} and Fig. \ref{fig:ratio}. In Fig. \ref{fig:hist}, the $y$-axis denotes the cumulative number of cases evaluated, while the $x$-axis denotes the computing time and the number of backups. We plot results for both environments $\mathcal{E}_1$ (on the left) and $\mathcal{E}_2$ (on the right). The red line shows computing time taken, while the blue line shows the number of backups, and the dotted line refers to the P-MDP (our baseline) while the bold line refers to the AP-MDP (our proposed model). For the corresponding number of cases on the $y$-axis, we can see the time taken or the number of backups, as plotted by the four lines. In both environments $\mathcal{E}_1$ and $\mathcal{E}_2$, the AP-MDP finds solutions with a shorter computing time and a smaller number of backups in the majority of cases. The size of environment $\mathcal{E}_2$ is much larger than $\mathcal{E}_1$, and it therefore takes longer computing time and more backups. It should be noted that AP-MDP perform significantly better than P-MDP given the benefits of abstraction in large states spaces. 

In Fig. \ref{fig:ratio}, to compare the efficiency of the two algorithms we plot the ratio (that is, AP-MDP to P-MDP) for the same metrics. For both computing time and number of backups, a ratio less than $1.0$ indicates that AP-MDP is more efficient than P-MDP. The y-axis shows the cumulative number of cases, while the $x$-axis shows the ratio of the computing time taken. For a corresponding ratio on the $x$-axis ($r=0.2$, for example) we can see the number of cases that had a ratio $< r$). Therefore, a line that solves a larger number of cases (out of $100$) at a smaller ratio is a better solution. The four lines refer to different environments when solved at different levels. For example, ($\mathcal{E}_1$, $l=1$) refers to the smaller environment at level $1$. In $\mathcal{E}_1$, AP-MDP is better in $72$ among the $100$ cases with respect to the computing time and for $71$ cases with respect to the number of backups. In $\mathcal{E}_2$, AP-MDP is better in $86$ among $100$ cases with respect to the computing time and for $89$ cases with respect to the number of backups. 
The AP-MDP decomposes the problem and therefore has to solve more MDPs than the P-MDP. This means that in certain cases, especially in the smaller environment where abstraction is unnecessary, this approach is not faster. However, in the larger environment, employing abstraction increases the efficiency by reducing the size of each problem. 
To clearly show the effect of abstraction, we run simulations with atomic propositions in higher levels ($AP^1$ and $AP^2$), to assess how much abstraction helps when dealing with high-level commands. In $\mathcal{E}_1$, the computing time ratio is less than $1.0$ in $95$ cases and the number of backups ratio is less than $1.0$ in $99$ cases. In the larger environment $\mathcal{E}_2$, the computing time ratio and the number of backups ratio are less than $1.0$ in all cases. 


\begin{figure}
\centering
\subfigure[First floor]{\includegraphics[width=.48\columnwidth]{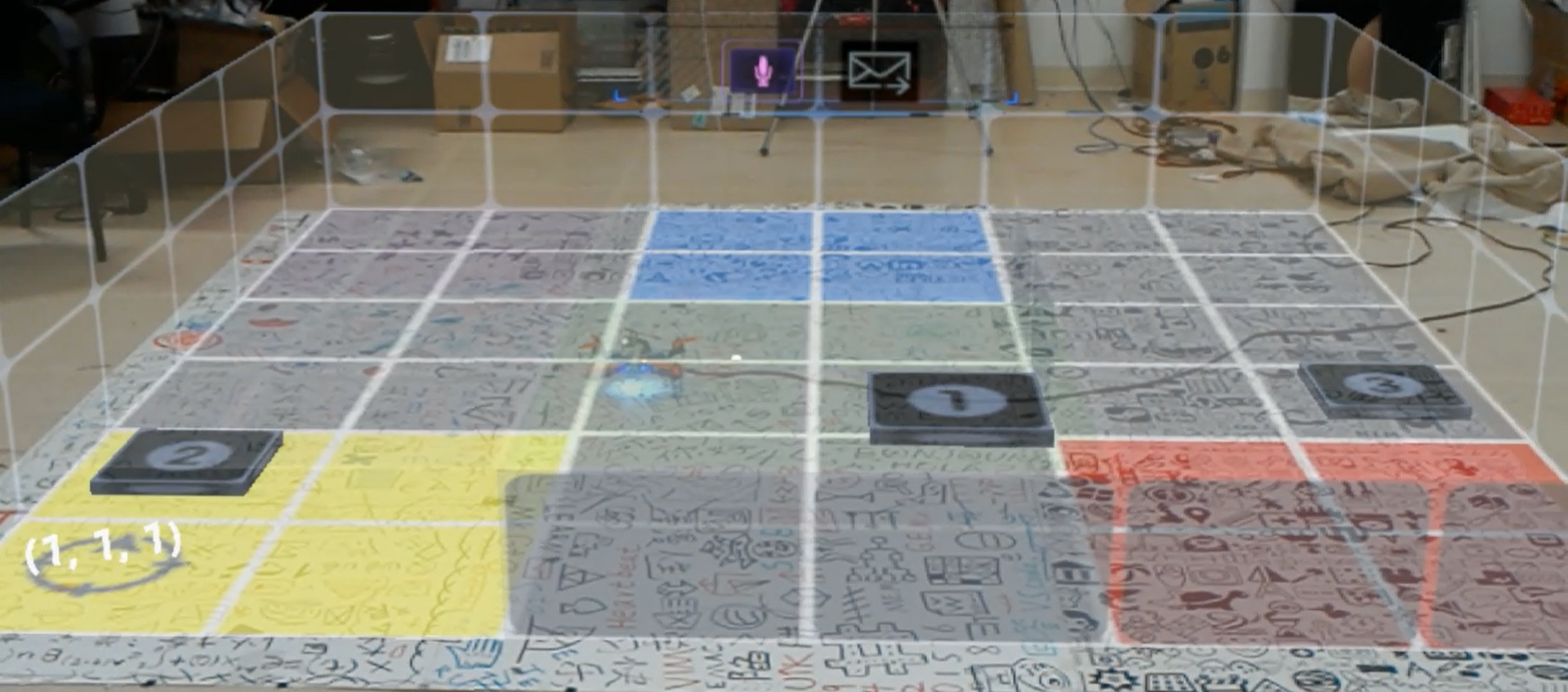}\label{fig:drone_env1}}
\subfigure[Second floor]{\includegraphics[width=.48\columnwidth]{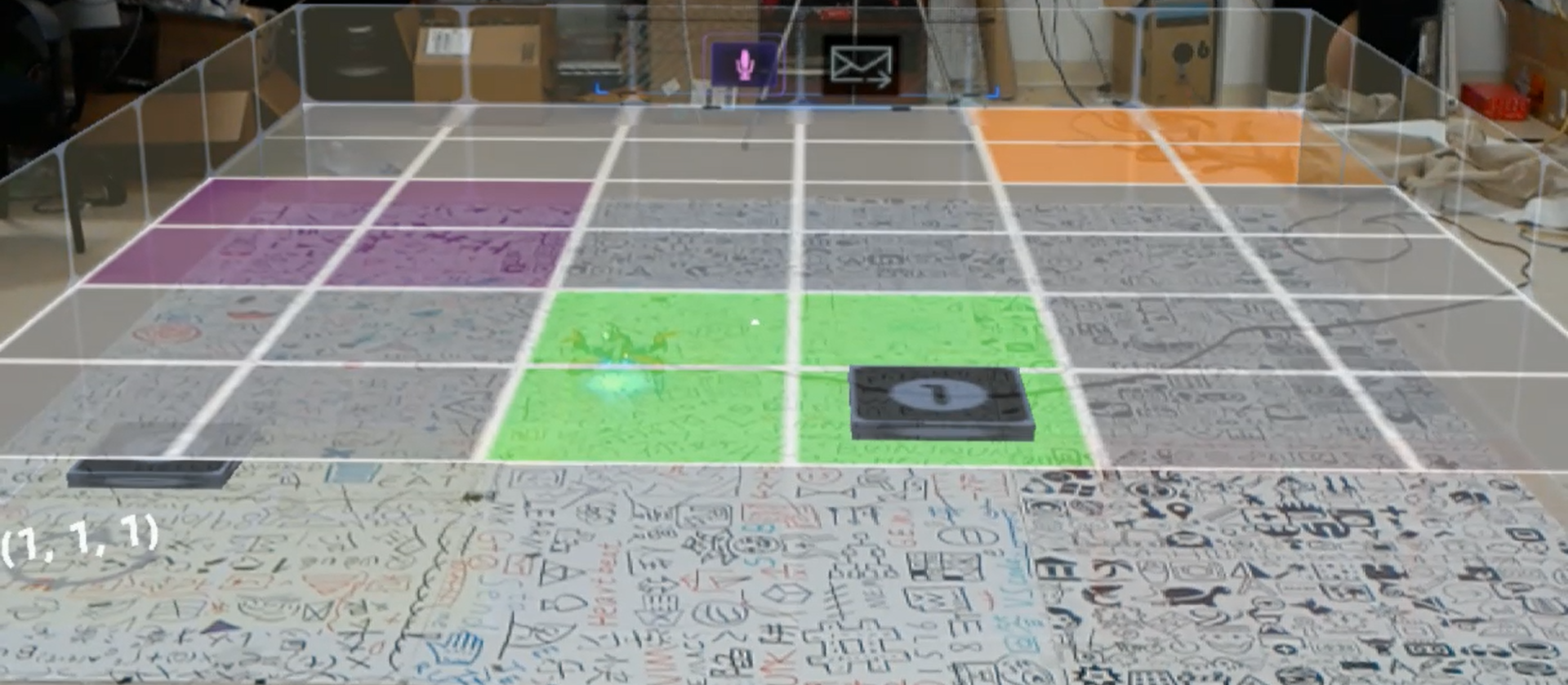}\label{fig:drone_env2}}
    \caption{Figures of the two-floor environment for our drone demonstrations as viewed through the HoloLens, taken from our video.
}\label{fig:drone_env}
\end{figure}

\subsection{Language grounding results}
We observe that the accuracy of the model drops on the held-out LTL commands. This problem of zero-shot generalization (specifically, the ability to generalize to samples unseen during training) has been widely studied \cite{lake2017still, koehn2017six, gopalan2018sequence} for neural sequence-to-sequence models that cannot handle compositionality and the ability of models to learn meaning representations for given natural language sentences \cite{DBLP:journals/corr/abs-1802-04302}. We also observe cases where  changes in word order affect the translated LTL output of the model. Consider the command \textit{``avoid the blue room until you go to landmark 1''}, ($\neg$\textit{blue\_room} $\U$ \textit{landmark\_1}) for example. Variations in our collected data include sentences like \textit{``until you go to landmark 1, always avoid the blue room''} that change the ordering of referent words (\textit{blue\_room} and \textit{landmark\_1}) which are occasionally confused, and mapped to incorrect expressions such as ($\neg$\textit{landmark\_1} $\U$ \textit{blue\_room}). 
However, in the drone demonstrations, the sequence-to-sequence model correctly translate the given language commands (converted from speech) into LTL task specifications that are then solved using our proposed method.

\subsection{Robot Experiments}\label{sec:robot}
In addition to the simulations described above, we also test our proposed method on a drone. The PiDrone \cite{brand2018pidrone} is a quadcopter drone that is equipped with one downward-facing infrared sensor with a maximum range of $60cm$ to measure the drone's altitude,  and one downward-facing camera for localization over a textured surface. The drone's flight space is a $3m\times3m$ surface. We divide the space into a grid-based environment, as shown in Fig. \ref{fig:drone_env}, consisting of $2$ floors, each with $9$ rooms, and each room is a square made up of $4$ cells (each cell is $50cm\times50cm$). The action space for the drone in the grid-based environment is (\textit{north, south, east, west, up, down}), where each action changes the drone's location by $1$ cell. We visualize the environment through mixed reality using a Microsoft HoloLens \cite{chen20153d}. Colored rooms and landmarks (boxes each with the size of $1$ cell) to aid path planning and specify goal positions were set up in a Unity3D virtual environment running on the HoloLens.

In our demonstration, the drone is given a natural language instruction through speech. This is converted using Google's speech-to-text, and then translated by our trained sequence-to-sequence model into an LTL formula to be solved by the AP-MDP framework in real time.
The action sequence output by AP-MDP for the LTL expression is then used for the drone's navigation. The natural language commands were: \textit{``Navigate to the red room", ``Avoid landmark two until you have been to the blue room", ``Move to the orange room then the purple room", ``Go to landmark three then go to the yellow room"}. Video recordings of the drone demonstrations can be found at \url{https://youtu.be/zjtMEGUmkd8}.
\section{Conclusion}
This paper introduces a novel approach to combine the handling of non-Markovian task specifications in large environments by grounding complex language to LTL expressions and then decomposing tasks within an abstraction hierarchy to plan efficiently at higher levels where possible. We show that planning with abstractions allows the robot to correctly reach the goal location more efficiently, in terms of computing time and backups required, in over $95\%$ of tasks in a small environment and over $99\%$ of tasks in a larger environment.
We also show that this method of abstraction can handle LTL task specifications. 
Moreover, we present the largest existing dataset of natural language commands mapped to LTL expressions at different levels of abstraction. We demonstrate our approach with a PiDrone that navigates to the goal location along a correct path when given a human-uttered command.

While the language grounding model works fairly well to translate language to LTL formulae, it cannot fully handle expressions unseen during training and cannot always deal with simple changes in word-ordering and variations in the language. 
Future work in this direction can explore compositional models that can handle a wide range of expressions by learning to compose subparts together and then execute the required actions. Future work in the hierarchical setup can explore models that go beyond fixed hierarchies and state abstractions. If the AMDP transition hierarchies can be learned with model-learning methods on the fly, this will enable generalization to unseen environments and the ability to handle and properly execute a plan for a wider range of commands.

\section{Acknowledgments}
The authors would like to thank Nakul Gopalan for his insightful comments and edits. This work is supported by the National Science Foundation under grant numbers IIS-1637614 and IIS-1652561, and the National Aeronautics and Space Administration under grant number NNX16AR61G. 

\bibliographystyle{IEEEtran}
\bibliography{IEEEabrv,ref}

\end{document}